\documentclass{article}

\usepackage{arxiv}

\usepackage[utf8]{inputenc} 
\usepackage[T1]{fontenc}    
\usepackage{hyperref}       
\usepackage{url}            
\usepackage{booktabs}       
\usepackage{amsfonts}       
\usepackage{nicefrac}       
\usepackage{microtype}      
\usepackage{lipsum}		
\usepackage{graphicx}
\usepackage{natbib}
\usepackage{doi}

\usepackage{xcolor}         

\usepackage{subfigure}

\usepackage{float}
\usepackage{tabularx}
\usepackage[fleqn]{amsmath}
\usepackage{algpseudocode}
\usepackage[]{algorithm2e}
\usepackage{amssymb}
\usepackage{hyperref}

\usepackage{wrapfig}
\usepackage{fancyhdr}
\usepackage{color}

\usepackage{eso-pic}
\usepackage{forloop}

\title{Approximate Bayesian Computation for Physical Inverse Modeling}

\author{%
  Neel Chatterjee\thanks{Corresponding author, chatt097@umn.edu} \\
  Department of Electrical and Computer Engineering\\
  University of Minnesota - Twin Cities\\
  Minneapolis, MN 55455 \\
  \texttt{chatt097@umn.edu} \\
   \And
   Somya Sharma \\
   Department of Computer Science and Engineering \\
  University of Minnesota - Twin Cities\\
  Minneapolis, MN 55455 \\
   \texttt{sharm636@umn.edu} \\
   \AND
   Sarah Swisher \\
   Department of Electrical and Computer Engineering \\
  University of Minnesota - Twin Cities\\
  Minneapolis, MN 55455 \\
   \texttt{sswisher@umn.edu} \\
   \And
   Snigdhansu Chatterjee \\
   School of Statistics \\
  University of Minnesota - Twin Cities\\
  Minneapolis, MN 55455 \\   \texttt{chatt019@umn.edu} \\
}

\date{}

\renewcommand{\shorttitle}{Approximate Bayesian Computation for Physical Inverse Modeling}

\hypersetup{
pdftitle={Approximate Bayesian Computation for Physical Inverse Modeling},
pdfauthor={Neel ~Chatterjee, Somya ~Sharma, Sarah ~Swisher, Snigdhansu ~Chatterjee},
pdfkeywords={approximate Bayesian computation, semiconductor model, gradient boosting, inverse model},
}

\begin{document}
\maketitle

\begin{abstract}
	Semiconductor device models are essential to understand the charge transport in thin film transistors (TFTs). Using these TFT models to draw inference involves estimating parameters used to fit to the experimental data. These experimental data can involve extracted charge carrier mobility or measured current. Estimating these parameters help us draw inferences about device performance. Fitting a TFT model for a given experimental data using the model parameters relies on manual fine tuning of multiple parameters by human experts. Several of these parameters may have confounding effects on the experimental data, making their individual effect extraction a non-intuitive process during manual tuning. To avoid this convoluted process, we propose a new method for automating the model parameter extraction process resulting in an accurate model fitting. In this work, model choice based approximate Bayesian computation (aBc) is used for generating the posterior distribution of the estimated parameters using observed mobility at various gate voltage values. Furthermore, it is shown that the extracted parameters can be accurately predicted from the mobility curves using gradient boosted trees. This work also provides a comparative analysis of the proposed framework with fine-tuned neural networks wherein the proposed framework is shown to perform better.
\end{abstract}


\section{Introduction}
\label{intro}

The transport characteristics of metal-oxide thin film transistors can be assessed by measuring the mobility as a function of gate voltage in the channel. Often after obtaining experimental data, the mobility - gate voltage curves can be reproduced via numerical and physical models \cite{hsieh2008modeling,bae2011analytical,hsieh2008p,chatterjee2021modeling}. However, these simulation models are parameterized and require manual tuning of parameters to replicate the real-world experimental data. While these mathematical models provide a generalized framework for testing the performance of particular materials, the manual fine tuning process can become intractable when a large number of devices are being assessed simultaneously. It is useful to have a methodology where the parameters, $X$, associated with a model need to be retrieved, given a mobility curve $Y$. In this case, the mobility is a function of a given physical/numerical model, $F$, which can be represented as $Y=F(X) +\eta$, where $\eta$ denotes the noise term with an unknown distribution.

Approximate Bayesian computation (aBc) encompasses likelihood-free methods, which only relies on a given prior distribution for the input parameters. Stochastic simulations are performed where the forward model is called repetitively for different input parameters. For each input parameter, a similarity measure to the observed data is computed and only those input parameters are accepted where the computed similarity measure is below a certain threshold \cite{csillery2010approximate,marjoram2003markov,sisson2018handbook,beaumont2019approximate}. The stricter the threshold, smaller the number of samples accepted \cite{sisson2018handbook}. This method allows us to approximate the posterior distribution for each given input parameter, given some observation (mobility, in this case). However, when the input dimension is high, aBc can be computationally expensive. To circumvent this problem, adaptive aBc techniques have been proposed where the proposal distribution is tuned sequentially to sample specific regions of the input parameter distributions more extensively \cite{beaumont2009adaptive,meeds2014gps}.

In these contexts, we apply a two-stage aBc coupled with gradient boosted (GB) trees to accurately predict the input parameters of any observed mobility curves. On both noise-free simulated datasets and real world experimental datasets, we demonstrate the applicability of our framework. The first stage aBc narrows down our search for parameters, while the second stage aBc refines the parameter space further. This produces a posterior distribution of parameters that could have given rise to the mobility curves. The aBc implementation can be seen as an adaption of rejection aBc where all generated samples are stored for further computation instead of solely relying on the accepted samples. However, to obtain precise point estimates, we implement a GB model to predict parameters from mobility curves. We also compare our results to neural network models. These results showcase that the inverse model has the potential to elude the need for the arduous and inefficient manual tuning of parameters.

\section{Methodology}
\label{Methodology}

\textit{\textbf{Approximate Bayesian Computation - Preliminary Search}}. Approximate Bayesian computation (aBc) was used to approximate the posteriordistribution of the parameters for the given data. For the preliminary estimation, prior distribution for parameters are set as uniform. Using Bayesian optimization, those parameter values are favored that result in minimizing the difference of TFT model based mobility estimation from the observed mobility curve. Over $N_{prelim}$ trials, the simulated mobility curve, $\hat{\mu} (V_g)$ is compared to the input data, $\mu (V_g)$ and the mean squared error (summary statistic) is computed between the two. If the loss ($\epsilon$) is below a given threshold, $\epsilon_0$, the parameters which resulted in the simulated mobility curve are accepted. Tree-structured Parzen estimator (TPE) is used to select the next set of parameters. 

\textit{\textbf{Approximate Bayesian Computation - Refined Search}}. Once we perform the preliminary estimation, we obtain the parameters, $\mu_{0,Pre}, T_{0,Pre}, D_{0,Pre}, N_{t,Pre}, \text{ and } E_{t,Pre}$, which yield the lowest MSE. Standard deviation using the posterior distribution is also computed, $\mu_{0,\sigma}, T_{0,\sigma}, D_{0,\sigma}, N_{t,\sigma}, \text{ and } E_{t,\sigma}$. Using these two values for each given parameter, we define a new prior distribution as search space as shown in Algorithm \ref{algo:ABC_Refined}. This process is repeated for a given number of iterations, $N_{refined}$. After the two runs, we obtain posterior distributions for the different parameters used in the TFT model. This allows for a probabilistic inference for each parameter in the model.

\textit{\textbf{Gradient Boosting}}. These aBc trials still fail to provide precise parameter values, which we obtain from a gradient boosted tree fit. Using the set of chosen parameters and the corresponding mobility curve, we train a gradient boosted tree (GB) model. The input to the GB model are the mobility curves whereas the associated parameters which gave rise to these curves serve as our GB model output. Once the model is trained, we query the model to predict the set of parameters which might have generated our experimental mobility curve. 

\textit{\textbf{Deep Neural Network - Baseline Models}}. We trained a shallow (3 layers - Shallow NN) and a deep neural network (29 layers - Deep NN) as baselines for comparison. To train the neural networks, we used our physical model to generate data using multiple combinations of input parameters. Gaussian Processes based hyperparameter optimization \cite{frazier2018tutorial} was used to find the optimal architecture. Once the networks were trained, we can query the model using mobility curves and the neural network provides point estimates for the multiple input parameters. The results of the two networks are compared to our inverse model as shown in Table 2.

\section{Experiment Details}

\subsection{Noise Free Data}

\begin{figure*}
    \centering
    \includegraphics[width=\textwidth]{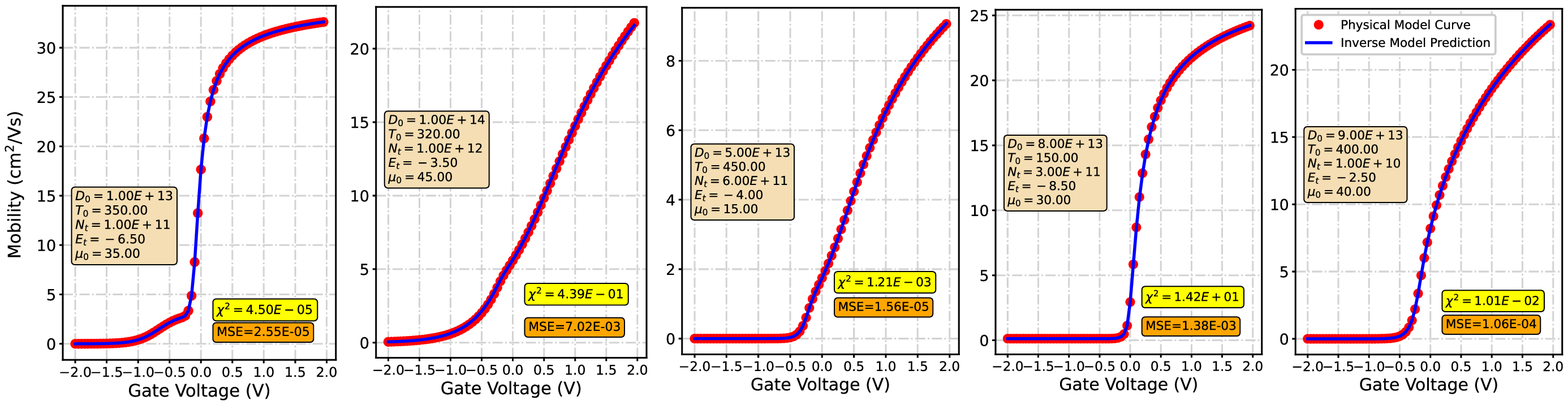}
    \caption{Experimental curves are shown in solid symbols. The parameters which were used to generate the curve are shown as insets in a text box. The predicted curves were simulated using the parameters estimated by our inverse model. The chi-squared statistic denoted the goodness of fit \textit{between the parameters} used to generate the mobility curve and the ones predicted by gradient boosting. The MSE denotes the mean squared error between the curve generated from the physical model using the parameters in the text box and the curve generated by the physical model using the parameters predicted by our inverse model.}
    \label{All_Mobility_Curves}
\end{figure*}

In the noise-free scenario, we use the physical model to simulate mobility curves using five different sets of parameters (Experiments 1 to 5). The parameter estimation for these curves is carried out by the approach outlined in section \ref{Methodology}. The range of the priors used at the beginning of the inverse model flow are detailed in Algorithm \ref{algo:ABC_Prelim}. Fig. \ref{All_Mobility_Curves} shows all the mobility curves simulated using the physical model. We use the parameters estimated by the GB trees to simulate the mobility curves shown in Fig. \ref{All_Mobility_Curves}. It can be observed that our inverse model flow is able to identify the parameters from which the mobility curve was generated. The chi-squared statistic is used as an evaluation metric and is computed using, $\chi^2=\sum \frac{(O_i-E_i)^2}{E_i}$, where $O_i$ are the observed parameters obtained from gradient boosting and $E_i$ are the parameters which was used to simulate the experimental curves. The baseline neural networks were not able to predict the correct set of input parameters which yield the lowest MSE with respect to the input mobility curve. The results for the five different experiments are shown in Table \ref{table:2}. 

\vspace{20pt}
\begin{minipage}{0.46\textwidth}

\rule{0.92\textwidth}{0.2pt}

\begin{algorithm}[H]
\small
\SetAlgoLined



\KwResult{$\mu (V_g)$ and $\mu_0,T_0,D_0,E_t,N_t$}

\textbf{Input}: $N_{prelim}$ and $\epsilon_0$ 


\textbf{Initialize:}

\begin{enumerate}
    \item array $\mu_{accepted}=[\hspace{15pt}]_{N \times 1}$
    \item array $Parameters = [\hspace{15pt}]_{N \times 5}$
\end{enumerate}

\textbf{Algorithm:}

\For{i in 1:N}{

$\epsilon_0=1$

\While{$\epsilon < \epsilon_0$}{

$\mu_0 \sim [1,50]$

$T_0 \sim [50,600]$

$D_0 \sim [1E12,2.28E14]$

$N_t \sim [1E10,1E13]$

$E_t \sim [-10,-3]$\\
$\hat{\mu}(V_g) = TFT_{model} (\mu_0,T_0,D_0,N_t,E_t)$

$\epsilon[i]=|\mu(V_g)-\hat{\mu}(V_g)|$

}

$\mu[i] = \hat{\mu}(V_g)$

$Parameters[i] = [\mu_0,T_0,D_0,N_t,E_t]$

}

\rule{0.92\textwidth}{0.2pt}

\caption{Approximate Bayesian Computation of Parameters - Preliminary Estimation}
\label{algo:ABC_Prelim}
\end{algorithm}

\rule{0.92\textwidth}{0.2pt}

\end{minipage}
\hfill
\begin{minipage}{0.46\textwidth}

\rule{0.92\textwidth}{0.2pt}

\begin{algorithm}[H]
\small
\SetAlgoLined



\KwResult{$\mu (V_g)$ and $\mu_0,T_0,D_0,E_t,N_t$}

\textbf{Input}: $N_{refined}$ and $\epsilon_0$ 


\textbf{Initialize:}

\begin{enumerate}
    \item array $\mu_{accepted}=[\hspace{15pt}]_{N \times 1}$
    \item array $Parameters = [\hspace{15pt}]_{N \times 5}$
\end{enumerate}

\textbf{Algorithm:}

\For{i in 1:N}{

$\epsilon_0=1$

\While{$\epsilon < \epsilon_0$}{

$\mu_0 \sim [\mu_{0,Pre}-\sigma,\mu_{0,Pre}+\sigma]$

$T_0 \sim [T_{0,Pre}-\sigma,T_{0,Pre}+\sigma]$

$D_0 \sim [D_{0,Pre}-\sigma,D_{0,Pre}+\sigma]$

$N_t \sim [N_{t,Pre}-\sigma,N_{t,Pre}+\sigma]$

$E_t \sim [E_{t,Pre}-\sigma,E_{t,Pre}+\sigma]$

$\hat{\mu}(V_g) = TFT_{model} (\mu_0,T_0,D_0,N_t,E_t)$

$\epsilon[i]=|\mu(V_g)-\hat{\mu}(V_g)|$

}

$\mu[i] = \hat{\mu}(V_g)$

$Parameters[i] = [\mu_0,T_0,D_0,N_t,E_t]$

}

\rule{0.92\textwidth}{0.2pt}

\caption{Approximate Bayesian Computation of Parameters - Refined Estimation}

\label{algo:ABC_Refined}
\end{algorithm}

\rule{0.92\textwidth}{0.2pt}

\end{minipage}

\vspace{20pt}

\begin{table}[H]

\begin{center}

\scriptsize

\caption{$\chi^2$ and MSE values for inverse model, shallow neural network and deep neural network over five different experiments.}

 \begin{tabular}{p{0.2\textwidth}>{\centering}p{0.2\textwidth}>{\centering}p{0.2\textwidth}>{\centering\arraybackslash}p{0.2\textwidth}}
 \toprule
 Metrics & Inverse Model & Shallow NN & Deep NN \\
 \midrule
 $\chi^2$ (Exp 1) & \textbf{4.5e-5} & 0.37 & 6.21 \\
 \hline
 MSE (Exp 1) & \textbf{2.5e-5} & 0.72 & 22.84 \\
 \hline
 $\chi^2$ (Exp 2) & 0.43 & \textbf{0.043} & 25.42 \\
 \hline
 MSE (Exp 2) & \textbf{0.0070} & 3.46 & 5.78 \\
 \hline
 $\chi^2$ (Exp 3) & \textbf{0.0012} & 0.88 & 2.44 \\
 \hline
 MSE (Exp 3) & \textbf{1.5e-5} & 0.05 & 13.80 \\
 \hline
 $\chi^2$ (Exp 4) & 14.21 & 8.69 & \textbf{8.05} \\
 \hline
 MSE (Exp 4) & \textbf{0.001} & 3.37 & 35.46 \\
 \hline
 $\chi^2$ (Exp 5) & \textbf{0.010} & 1.15 & 7.04 \\
 \hline
 MSE (Exp 5) & \textbf{1e-5} & 5.01 & 0.45 \\
 \hline
 $\chi^2$ mean ($\pm$ S.D.) & 2.93 ($\pm$ 5.6) & \textbf{2.23 ($\pm$ 3.2)} & 9.84 ($\pm$ 8.0) \\
 \hline
 MSE mean ($\pm$ S.D.) & \textbf{1.71e-3 ($\pm$ 2e-3)} & 2.53 ($\pm$ 1.85) & 1.57 ($\pm$ 12.4) \\
 \bottomrule

\end{tabular}
\label{table:2}
\end{center}
\end{table} 

\subsection{Validation with Experimental Data}

\begin{figure*}
    \centering
    \includegraphics[width=0.80\textwidth]{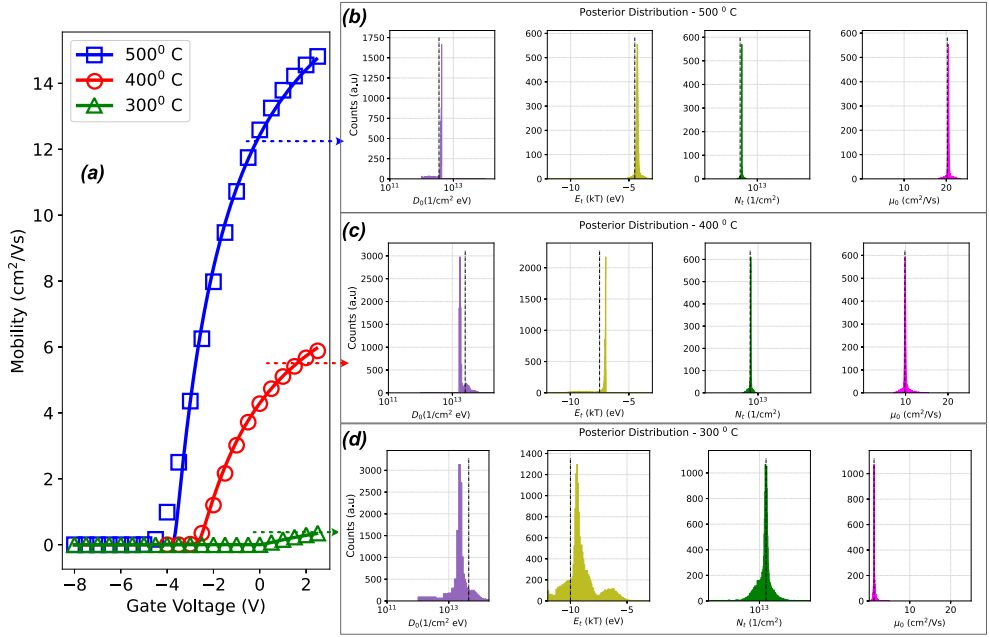}
    \caption{ (a) Input parameters extracted for mobility curves obtained from experiments. The markers are the data obtained experimentally from \cite{chatterjee2021modeling} and the solid lines are simulated curves using our physical model. (b), (c) and (d) show the posterior distributions obtained for devices processed at 500 , 400 and 300 $^0$C respectively, obtained after the two-stage aBc search for the input parameters. The black dashed lines in the three plots show the parameters estimated by the gradient boosted trees after the two stage aBc, which were used to simulate the mobility curves (solid lines) in (a).}
    \label{All_Mobs_Posterior}
\end{figure*}

We also demonstrate the applicability of the inverse model on real world data to find the parameters for experimental data obtained from fabricated devices. We obtain the data from indium zinc oxide (IZO) TFTs which is a candidate material for metal oxide TFTs \cite{chatterjee2021modeling}. The posterior distributions of the input parameters across the three devices are shown in Fig \ref{All_Mobs_Posterior}. We can observe that $D_0$ and $N_t$ decrease as we go from 300 to 500 $^0$C post-processing temperature. The parameter $E_t$ becomes less negative and $\mu_0$ increases by an order of magnitude from the 300 to 500 $^0$C device. Similar trends were observed previously \cite{chatterjee2021modeling}. Along with the posterior distributions, the experimental mobility curves and the simulated mobility curves are also shown in Fig. \ref{All_Mobs_Posterior}(a). To generate the simulated mobility curves, the parameters predicted by the GB trees at the end of the two-stage aBc were used. We obtain an excellent fit between the data obtained from experimental real-world devices and the ones predicted by our inverse model.    

\section{Conclusion and Future Work}

In this work, we present a general method to explore the search space of an inverse problem using aBc coupled with a GB trees at the end, which accurately predicts the input parameters. We validated this approach with five different experiments consisting of simulated data where we obtained reasonable chi-squared values with respect to the input parameters while obtaining low MSE values with respect to the observed mobility curve. We ultimately used our inverse model to infer parameters for experimental data. It was observed that our inverse model allowed us to draw some useful conclusions, which are similar to the ones observed previously \cite{chatterjee2021modeling}. We are currently exploring how to incorporate this technique in other physical problems to infer the input parameters. Since, TFT model parameter posterior distribution does not exhibit multi-modality, using a deterministic modeling method to obtain point estimates is appropriate in our application. For applications where multi-modality is a concern, probabilistic modeling methods can be explored. Also, using neural networks with standard backpropagation in these cases seem inefficient since the input values can take a multitude of values, which is never known beforehand. Training data incorporating the whole range of possible values will need to be generated first to train a neural network. The granularity of the data that can be generated is then limited by the computation of the physical model, which can be quite expensive. Using physics informed neural networks (PINN) to reduce the need of a large training dataset is a viable alternative, one we are currently exploring.  

\section{Broader Impact}

In our work, we successfully estimate parameters for accurately estimating mobility curves. This creates opportunity to efficiently create general frameworks for testing performance of different materials in device fabrication. For instance, new materials for fabricating high-performing thin-film transistors (TFTs) for display applications can be discovered and optimized. As such, the methodology can potentially be applied for parameter estimation in any model involving physical processes and further investigation in that respect is required. We believe that this work does not have any foreseeable negative societal consequence.

\bibliographystyle{unsrtnat}
\bibliography{template}


\end{document}